%% file: main.tex
\documentclass[sigconf,screen,balance]{acmart}

\AtBeginDocument{%
  }

\setcopyright{acmcopyright}
\copyrightyear{2018}
\acmYear{2018}
\acmDOI{XXXXXXX.XXXXXXX}

\acmConference[MM '23]{Ottawa '23: ACM MULTIMEDIA}{October 29-- November 2, 2023}{Ottawa, Ontario, Canada}
\acmPrice{15.00}
\acmISBN{978-1-4503-XXXX-X/18/06}

\input{math}

\usepackage{graphicx}
\usepackage{amsmath}
\usepackage{times}
\usepackage{epsfig}
\usepackage{multirow, float, booktabs, boldline, hhline}
\usepackage{algorithm}
\usepackage{algpseudocode}
\usepackage{caption}
\usepackage{subcaption}
\usepackage{ctable}
\usepackage{diagbox}
\usepackage{cellspace}
\usepackage{color, colortbl}
\usepackage{multirow}
\usepackage{dsfont}
\usepackage{balance}
\usepackage{enumitem}
\usepackage{balance}

\acmSubmissionID{570}

\begin{document}

\title{Zero-Shot Learning by Harnessing Adversarial Samples}

\author{Zhi Chen$^{1}$, Pengfei Zhang$^{1}$, Jingjing Li$^{2}$, Sen Wang$^{1}$, Zi Huang$^{1}$}
\affiliation{
  \institution{$^1$School of Electrical Engineering and Computer Science, The University of Queensland, Australia}
  \country{}
}
\affiliation{
  \institution{$^2$School of Computer Science and Engineering, University of Electronic Science and Technology of China}
  \country{}
}
\email{ {zhi.chen, pengfei.zhang, sen.wang}@uq.edu.au, lijin117@yeah.net, huang@itee.uq.edu.au}

\renewcommand{\shortauthors}{Chen et al.}

\begin{abstract}

Zero-Shot Learning (ZSL) aims to recognize unseen classes by generalizing the knowledge, \textit{i.e.,} visual and semantic relationships, obtained from seen classes, where image augmentation techniques are commonly applied to improve the generalization ability of a model. However, this approach can also cause adverse effects on ZSL since the conventional augmentation techniques that solely depend on single-label supervision is not able to maintain semantic information and result in the \textit{semantic distortion} issue consequently. In other words, image argumentation may falsify the semantic (e.g., attribute) information of an image. To take the advantage of image augmentations while mitigating the semantic distortion issue, we propose a novel ZSL approach by Harnessing Adversarial Samples (HAS). HAS advances ZSL through adversarial training which takes into account three crucial aspects: \textbf{(1) robust generation} by enforcing augmentations to be similar to negative classes, while maintaining correct labels, \textbf{(2) reliable generation} by introducing a latent space constraint to avert significant deviations from the original data manifold, and \textbf{(3) diverse generation} by incorporating attribute-based perturbation by adjusting images according to each semantic attribute's  localization. Through comprehensive experiments on three prominent zero-shot benchmark datasets, we demonstrate the effectiveness of our adversarial samples approach in both ZSL and Generalized Zero-Shot Learning (GZSL) scenarios. Our source code is available at \color{red}{\url{https://github.com/uqzhichen/HASZSL}}.

\end{abstract}

\ccsdesc[500]{Computing methodologies ~ Computer vision}

\keywords{zero-shot learning, adversarial training}


\maketitle

\section{Introduction}

Traditional image recognition systems operate within a predefined label space, predicting categories based on a predetermined set of labels. To expand beyond this limited label space and accommodate emerging labels, one option is to laboriously retrain the recognition model using data associated with the new labels. Zero-Shot Learning (ZSL) \cite{xian2017zero,kodirov2017semantic} offers an alternative solution, presenting a flexible and scalable approach to incorporate new classes into the model without requiring retraining. This adaptability allows ZSL to efficiently accommodate novel categories, making it a more effective solution for expanding the capabilities of image recognition systems.

ZSL entails associating visual patterns with semantic vectors, which are typically annotated attributes. These attributes consist of multiple dimensions, each representing a semantic feature of the objects being analyzed. For example, one dimension could represent the "breast color" of a bird, with the value being "red". Many research efforts \cite{xu2020attribute,liu2021goal,chen2022msdn} have enabled ZSL methods to exhibit a strong ability to localize, effectively allowing them to attend to the regions of the images that are most relevant to the semantic meanings in the attributes. This capability of ZSL methods to localize and attend to the relevant regions of the images is a key factor in their improved accuracy and effectiveness. 

\begin{figure}[t]
    \centering
    \includegraphics[width=1\linewidth]{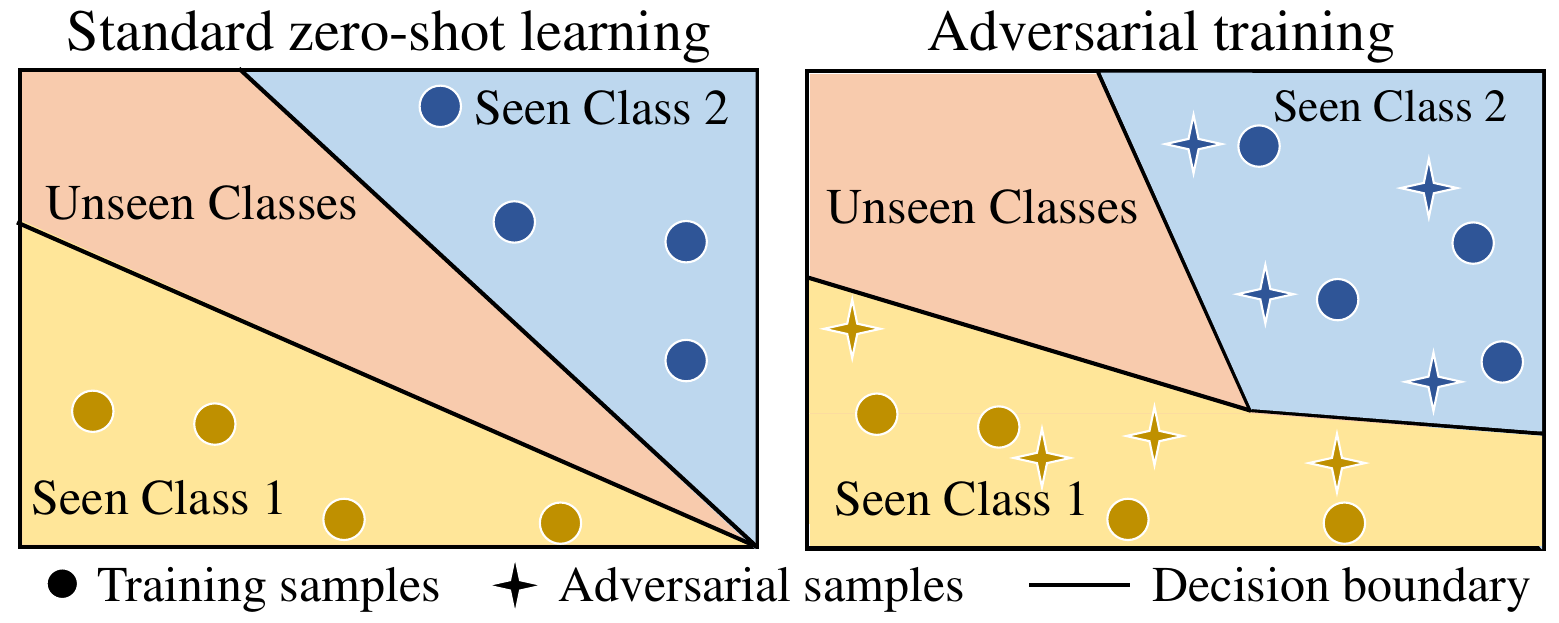}
    \caption{Our method employs adversarial training to direct the ZSL model towards generating augmentations that are conditioned on attributes to diversify the training images.}
    \vspace{-10pt}
    \label{fig:intro}
\end{figure}

\begin{figure*}[t]
    \centering
    \includegraphics[width=1\linewidth]{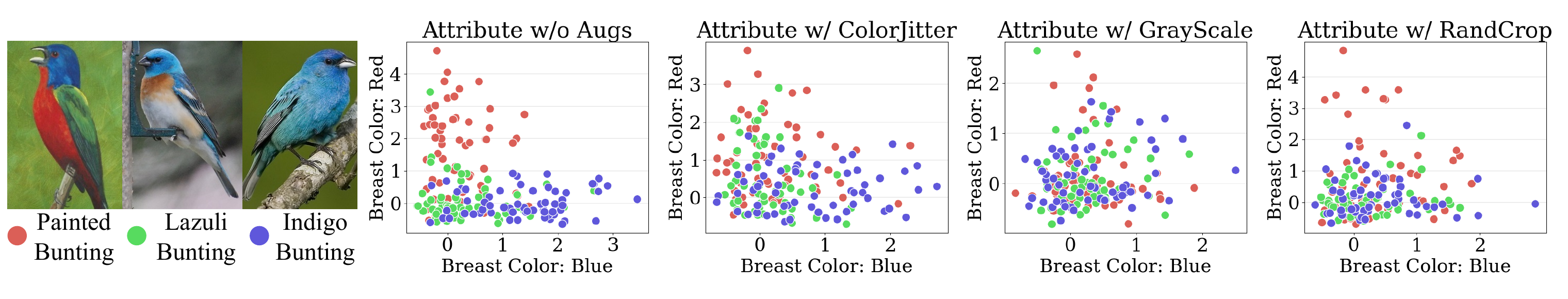}
    \vspace{-10pt}
    \caption{A close look at the semantic distortion problem. We observe that traditional image augmentation techniques lead to confusion in the ZSL model, causing it to generate incorrect attributes. This, in turn, results in indistinguishable attribute space.}
    \vspace{-5pt}
    \label{fig:aug_vis}
\end{figure*}

ZSL is fundamentally an image recognition task, where image augmentation techniques are commonly utilized to enhance the generalization ability. However, we observe that image augmentations can cause detrimental effects on ZSL. As illustrated in Figure \ref{fig:aug_vis}, we deduce two attributes (e.g., Blue Breast and Red Breast) for three bird categories using the ZSL model APN \cite{xu2020attribute}. The model accurately predicts the breast color of the three bird categories without augmentation. Nevertheless, when image augmentations are applied to raw images, the model produces incorrect values for the two attributes, causing the categories to become indistinguishable. We have also conducted a series of experiments, involving various types of image augmentation techniques in training a ZSL model, quantitatively observing performance drop. We term the problem caused by direct pixel manipulation on images as \textbf{\textit{semantic distortion}}. The causes of this problem can be attributed to two main factors: (1) Conventional image augmentation applies a set of \textit{label-preserving} transformations, which, despite improving generalization ability, can introduce semantic distortion issues due to their lack of \textit{semantic preservation}. This is because these data augmentation strategies \cite{shorten2019survey, zhong2020random,taylor2018improving} apply direct pixel manipulation on images, which can modify the semantic essence of the data, conflicting with ZSL objectives. Unlike supervised learning, which relies on a single label for guidance, ZSL demands dense semantics as strong supervision, making aggressive image augmentation susceptible to creating semantic ambiguity and compromising the semantic coherence prioritized in ZSL.  (2) An intriguing property of neural networks is that the entire space of activations, rather than individual units, contains the bulk of semantic information \cite{DBLP:journals/corr/SzegedyZSBEGF13}, making the entire space less sensitive to pixel manipulation. In ZSL, however, one can inspect the final individual units (i.e., attribute prediction), which usually attend to regions containing a specific semantic meaning \cite{xu2020attribute,liu2021goal}. Consequently, subtle pixel changes can lead to incorrect predictions in individual units. Based on these limitations, exploring alternative methods for enhancing ZSL performance without compromising semantic coherence is essential.

In light of the controllable image manipulation achieved by adversarial training \cite{DBLP:journals/corr/GoodfellowSS14,xie2020adversarial,herrmann2022pyramid}, we resort to learning adversarial samples as augmentations to diversify the visual space without causing semantic distortion problem, as shown in Figure \ref{fig:intro}. However, we identify three critical challenges of learning adversarial samples for ZSL. First, conventional adversarial training creates adversarial samples by maximizing the classification cross-entropy loss, which is an explicit way of perturbing images to confuse the model to make wrong predictions. However, the goal of generating adversarial samples for zero-shot learning is not to leverage the adversarial samples to confuse the model, but rather to facilitate the generalization ability on diverse images. Second, the learned visual features correspond to the predicted attributes. The latent space requires explicit constraint to prevent significant shifts from the original space while perturbing the original images. Third, ZSL models exhibit strong localization ability, but how to leverage the localization ability when generating adversarial samples remains unclear. To this end, we propose a novel ZSL approach by Harnessing Adversarial Samples (HAS). Specifically, to address the first robustness issue, bearing in mind that we still want to ensure the model recognizes the adversarial samples, but more importantly, perceives the differences, we propose to make the images visually similar to negative classes based on the model's understanding of differences among classes. This is achieved by maximizing the entropy of the classification probability so that the model could assign higher probabilities to other classes. To ensure the model recognizes the adversarial samples, the classification loss still needs to be minimized, so that the model could still perceive the adversarial images as correct ones. Second, to prevent significant shifts from the original space, we use the visual features of the original images to constrain the adversarial learning, so that the learned adversarial samples cannot produce significantly different visual features that cause the semantic distortion problem. Lastly, considering the attention maps on the last layer exhibit the localization probability of different attributes, we propose to explicitly perturb the attention maps by making the model give weak responses to the attribute regions by applying entropy loss and minimizing the total probability magnitude. Through iterative training with the adversarial samples, we can enhance the localization ability of a ZSL model.
Through comprehensive experiments on three prominent zero-shot benchmark datasets, we demonstrate the effectiveness of our adversarial samples approach in both ZSL and GZSL contexts. Our results highlight the ability of HAS in overcoming the limitations associated with traditional augmentation techniques within the ZSL domain. Furthermore, the proposed method provides a more coherent approach to image augmentation, which preserves the semantic information of the data while enhancing the model's generalization ability. 
The contribution of this work can be summarized as follows:
\vspace{-12pt}
\begin{itemize}[align=parleft,left=0em]
\item{We introduce a novel approach, Harnessing Adversarial Samples (HAS), to improve the generalization ability of ZSL models. HAS generates adversarial samples through controllable image perturbation, addressing the semantic distortion issue found in traditional image augmentations.}
\item{We consider three aspects in learning adversarial samples for ZSL. 1. Our method facilitates \textbf{robust generation} by maximizing the entropy of classification probabilities while minimizing cross-entropy loss, making images visually similar to negative classes without compromising model recognition. 2. We achieve \textbf{reliable generation} by utilizing the visual features of original images to stabilize adversarial learning, preventing significant shifts from the original visual space that could cause semantic distortion. 3. \textbf{Diverse generation} is promoted by leveraging the ZSL model's localization ability to perturb attention maps, enhancing the model's ability to focus on attribute regions.}
\item{Through comprehensive experimental study, we demonstrate that adversarial samples with proper learning objectives can effectively improve the generalization ability of ZSL models.}
\end{itemize}
\vspace{-2pt}
\newpage
\section{Related Work}
\subsection{Zero-Shot Learning} 
The ZSL task \cite{romera2015embarrassingly,xian2017zero,kodirov2017semantic,xian2018feature,chen2021semantics} consists in generalizing the knowledge learned from seen classes to unseen classes. With this goal, embedding-based methods learn a visual-to-semantic regression function by transforming the visual features to semantic information, \textit{e.g.,} attributes\cite{farhadi2009describing}, documents\cite{elhoseiny2013write}, w2v embeddings\cite{norouzi2013zero}. Zhu \textit{et al.} \cite{zhu2019semantic} proposed a Semantic-Guided Multi-Attention localization model (SGMA) that learns to discover the local features of the discriminative patches, which demonstrates strong localization performance with human annotations. The multi-attention loss encourages compact and diverse attention distribution by applying geometric constraints over attention maps. Xu \textit{et al.} \cite{xu2020attribute} proposed an Attribute Prototype Network (APN) that learns prototypes for each of the attributes. The learnable attribute prototypes are used as the 1×1 filter to convolve the feature maps with the feature maps from ResNet101 in order to attend to the region of interest.  As an extension to APN \cite{xu2020attribute}, to learn local features, Liu \textit{et al.} \cite{liu2021goal} proposed using the GloVe \cite{DBLP:conf/emnlp/PenningtonSM14} model to learn local features and to extract semantic vectors from the attribute names (\textit{e.g.,} "plain head"), which further represent the attribute prototypes. Chen \textit{et al.} \cite{chen2022msdn} proposed a mutually semantic distillation network to progressively distill the intrinsic semantic representations between and visual and attribute features. The other mainstream ZSL direction leverages generative models to hallucinate unseen visual features conditioned on the semantic information \cite{xian2018feature,yu2020episode,chen2021semantics,chen2020rethinking,chen2021mitigating,chen2020canzsl}. There are various generative models explored in this area, including GANs \cite{xian2018feature,zhu2018generative,su2022distinguishing}, VAEs \cite{wang2018zero,schonfeld2019generalized}, Flows \cite{shen2020invertible,chengsmflow}. 
Despite the success of image augmentation in supervised learning, its effectiveness in ZSL is still an open research problem. ZSL requires explicit supervision, \textit{i.e.,} attributes, which are very sensitive to subtle visual changes. In this context, we propose a novel approach that leverages adversarial samples as augmented images to improve ZSL performance. 

\vspace{-10pt}
\subsection{Adversarial Samples}
Adversarial samples, formed by adding imperceptible perturbations to images, can lead models to make wrong predictions. Attacks by adversarial samples \cite{DBLP:journals/corr/SzegedyZSBEGF13,dong2018boosting} pose a security concern in the field of recognition tasks. 
For generating adversarial examples in deep learning models, Fast Gradient Sign Method (FGSM) \cite{DBLP:journals/corr/GoodfellowSS14} is a simple and effective method. The paper provides a detailed explanation of FGSM and its applications, and it is considered a foundational work in the field of adversarial machine learning. The Worrisome is the phenomenon of adversarial examples \cite{biggio2013evasion}, imperceptibly perturbed natural inputs that induce erroneous predictions in state-of-the-art classifiers.
While early arts \cite{DBLP:conf/iclr/TsiprasSETM19,raghunathan2020understanding} suggest that there is a tradeoff between the robustness and the accuracy in recognition systems, there are works \cite{xie2020adversarial,herrmann2022pyramid} have shown that adversarial samples can also bring performance improvement. Our method further validates that adversarial features are also beneficial for ZSL recognition models, which agree with the conclusions drawn from these aforementioned studies.
Shafiee \textit{et al.} \cite{shafiee2022zero} proposed Attribute-based Universal Perturbation Generator to generate adversarial samples to attack zero-shot learning models. While our work focuses on leveraging adversarial samples to improve the zero-shot performance.


\section{Methodology}
\begin{figure*}[t]
    \centering
    \includegraphics[width=0.95\linewidth]{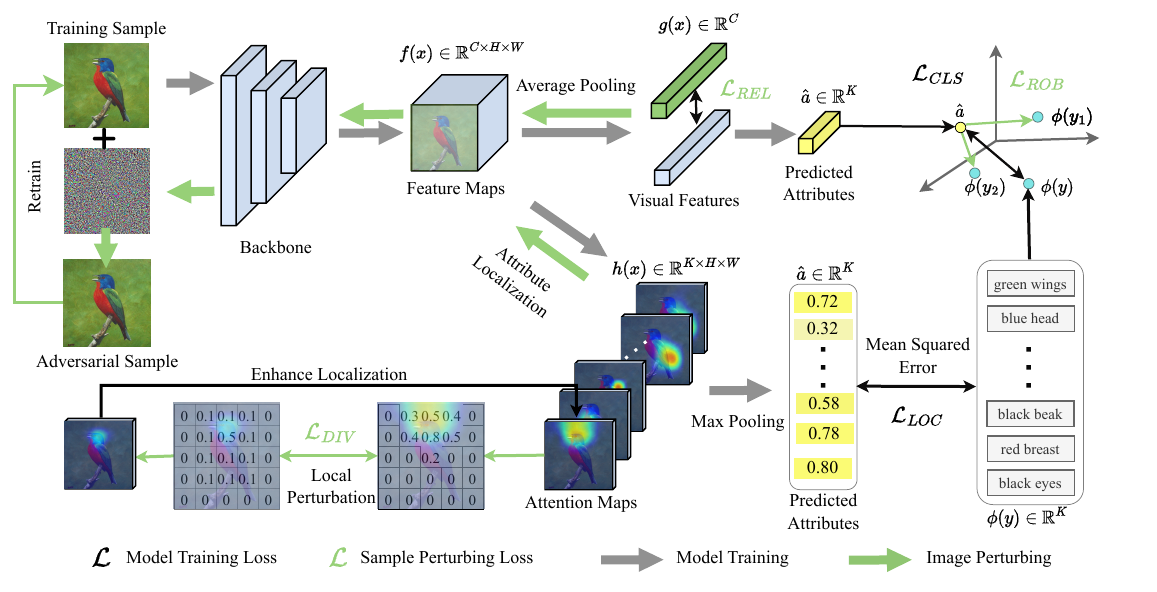}
    \vspace{-15pt}
    \caption{An illustration of our proposed HAS, which leverages adversarial training to guide the ZSL model in creating attribute-conditioned augmentations to  diversify the visual space. Using the APN baseline method \cite{xu2020attribute} for standard ZSL training, global visual features are projected into the attribute space for classification, and attribute attention maps enhance localization ability. Combining adversarial training with the baseline method, we generate attribute-conditioned image augmentations, resulting in a robust ZSL model that generalizes better to unseen classes while maintaining performance on known classes.}
    \label{fig:architecture}
    \vspace{-10pt}
\end{figure*}

\subsection{Preliminaries}
Assume a dataset $\mathcal{D}$ could be divided into two sets of classes $\mathcal{D}^s$ and $\mathcal{D}^u$, representing the seen classes and the unseen classes. The seen classes are used for training the ZSL model, i.e., $\mathcal{D}^s = \{I, y, \phi(y) | I \in \mathcal{I}, y \in \mathcal{Y}^s \}$ from seen classes $\mathcal{Y}^s$, where $I$ is an image in the RGB image space $\mathcal{I}$, y is the associated class label and $\phi(y) \in \mathbb{R}^K$ is the class-level semantic embedding annotated with K different semantic attributes. For the unseen classes $\mathcal{D}^u$, the class samples are unknown during training. 
The training procedure consists of two parts, i.e., standard ZSL training and adversarial training.
In standard training, we update the weights of the model to learn to predict the attributes of a given image and further infer the corresponding class, either from seen or unseen classes. In the adversarial training, we fix the model weights and update the training images via gradient ascent so that the adversarial samples can be yielded. The adversarial samples are then fed into the model for training with legitimate samples.

\subsection{Learning ZSL Model}
\label{stdZSLtraining}
We follow the baseline method Attribute Prototype Network (APN) \cite{xu2020attribute} to train a standard ZSL model.
Given an image $I$, the ResNet backbone produces the feature maps  $x = f(I) \in \mathbb{R}^ { C \times H \times W}$, where $C$, $H$ and $W$ represent the channel, height, and width respectively. The average pooling is performed on the feature maps to generate dense features $g(x) \in \mathbb{R}^{C}$. Then, a linear transformation $V \in \mathbb{R}^{C \times K}$ will be applied to predict the attribute vector $\hat{a} = g(x)^T V$. The classification loss function is then formulated as:
\begin{equation}
\begin{gathered}
\mathcal{L}_{CLS} = - \log \frac{\exp ( g(x)^T V \phi(y) ) }{ \sum_{ \hat{y} \in \mathcal{Y}^s} \exp (g(x)^T V \phi (\hat{y})) }.
\label{eq:cls}
\end{gathered}
\end{equation}

To further improve the localization ability of the ZSL model, a convolutional layer $ CV \in \mathbb{R}^{C \times K}$ is performed on the feature maps $\vx$. The filter size in the kernel is set to $1 \times 1 $. This operation results in the attribute attention map $h(x) \in \mathbb{R}^{K\times H \times W} $.  This operation localizes the semantic information of each attribute on each attention map. A max pooling operation is further applied on the attribute attention map to generate the attribute vector $\hat{a} \in \mathbb{R}^K$. Lastly, the attention localization loss can be formulated as:
\begin{equation}
\begin{gathered}
\mathcal{L}_{LOC} = \lVert \hat{a} - \phi(y)  \rVert_{2}^{2},
\label{eq:loc}
\end{gathered}
\end{equation}
where the mean squared error is calculated between the predicted attribute vectors and the ground truth attributes.



\subsection{Generating Adversarial Augmentations}
As discussed in the introduction section, the \textit{semantic distortion} problem poses a challenge in diversifying the visual space with traditional image augmentation techniques. 
Thus, we aim to devise a \textit{semantic-preserving} augmentation strategy, which diversifies the visual space while avoiding causing the semantic distortion problem. 

Inspired by Fast Gradient Sign Method (FGSM) \cite{DBLP:journals/corr/GoodfellowSS14}, learning perturbation on images creates adversarial samples. We develop a semantic-preserving algorithm that performs controllable image augmentations by perturbing images with gradient descent. However, adversarial samples created by FGSM aim to confuse the model to make wrong predictions. In contrast, in our method, the goal is to allow the model to conditionally diversify the images while the image should maintain the attribute prediction accuracy.  Our method does not involve additional parameters in the model for generating adversarial samples. Instead, the model introduced in Section \ref{stdZSLtraining} is used to provide guidance with different objective functions to update the input images. The overall loss function is formulated as follows:
\begin{equation}
\begin{gathered}
\mathcal{L}_{HAS} = \underbrace{\mathcal{L}_{CLS} - \lambda_1\mathcal{L}_{ROB}}_{Robustness} +~ \underbrace{\lambda_2\mathcal{L}_{REL}}_{Reliability} -~\underbrace{\lambda_3\mathcal{L}_{DIV}}_{Diversity},
\label{eq:HAS}
\end{gathered}
\end{equation}
where $\mathcal{L}_{CLS}$ and $\mathcal{L}_{ROB}$ ensure robust generation by maintaining correct label information,  $\mathcal{L}_{REL}$ facilitates reliable generation by preventing the images from semantic distortion, and $\mathcal{L}_{DIV}$ promotes diverse generation by perturbing the local attention maps.

Given the objective function $\mathcal{L}_{HAS}$, we iteratively generate the adversarial samples $I^{ad}$ with FGSM:
\begin{equation}
\begin{gathered}
I^{ad}_{0} = I, ~ I^{ad}_{t+1} = I^{ad}_{t} + \epsilon \sign \nabla_{I^{ad}_{t}} \mathcal{L}_{HAS}(\mW, I^{ad}_{t}, \phi(y)),
\label{eq:update}
\end{gathered}
\end{equation}
where $\epsilon$ is the perturbation strength, $\mW$ is the overall model parameters depicted in Section \ref{stdZSLtraining}, $\sign$ function applies a max-norm constraint on the gradients.

\subsection{Robust Generation}
To generate diverse adversarial samples without causing the model to misclassify, we propose to perturb the images to appear similar to negative classes. To do so, we resort to entropy maximization on the classification probabilities. The model's output probabilities, obtained after the Softmax layer, represent the confidence level for the predicted classes. By maintaining the highest confidence for the correct class while maximizing probability entropy across all classes, we enforce the model to update the image until it visually resembles other classes. The loss function that maintains robustness is formulated as follows:
\begin{equation}
\begin{gathered}
\mathcal{L}_{ROB} = - \Sigma_{{y} \in \mathcal{Y}^s}             p_y \log (p_y), ~
p_y = \frac{\exp ( g(x)^T V \phi(y) ) }{ \sum_{ \hat{y} \in \mathcal{Y}^s} \exp (g(x)^T V \phi (\hat{y})) }
\label{eq:div}
\end{gathered}
\end{equation}
where $p_y$ represents the probability of assigning the sample to class $y$. When generating adversarial samples for attacking a model \cite{DBLP:journals/corr/GoodfellowSS14}, it is common to apply gradient ascent on the images with classification loss, \textit{e.g.,} cross-entropy loss. This learning objective optimizes the images to mislead the model into making incorrect predictions. However, in our work, we aim to gently adjust the decision boundary in response to the visual perturbation, rather than provoke erroneous predictions. To achieve this, we still need to minimize the classification loss $\mathcal{L}_{CLS}$, which helps avoid drastic visual changes that might confuse the model and lead to improper boundary adjustments. By focusing on this objective, our approach promotes the generation of diverse adversarial samples that appear similar to negative classes while preserving the model's ability to classify correctly. This enables us to explore the model's sensitivity to visual perturbations and refine its decision boundaries without compromising its overall performance.

\subsection{Reliable Generation}
To further facilitate reliable perturbations applied to the images, we address the drift problem occurring in the latent space. The drift problem refers to a situation where an image produces the expected class probabilities according to the supervision provided, but its learned representations in the latent space undergo significant changes. Such changes affect the model's ability to generalize effectively and lead to instability in the learning process. To mitigate this issue, we introduce a constraint that aims to prevent the learned representations from deviating too much from the original ones. By doing so, we ensure that the latent features remain reliable and consistent. This constraint helps the model maintain a more robust and meaningful representation of the data in the latent space, which can enhance its overall performance and ability to generalize to unseen classes. The loss function that facilitates reliability can be formulated as:
\begin{equation}
\begin{gathered}
\mathcal{L}_{REL} = \lVert g(f(I)) - g(f(I^{ad}_{t})) \rVert_{2}^{2}
\label{eq:mea}
\end{gathered}
\end{equation}
where $g(f(I))$ represents the dense visual features generated by the model, and $g(f(I^{ad}_{t}))$ represents that of the perturbed images.


\begin{algorithm}[t]
\raggedright \textbf{Input}: seen dataset $\mathcal{D}^s$, training epoch $E$, batch size $N^b$, learning rate $\eta$
\\
\raggedright \textbf{Initialize}: Model weights $\mW$
\caption{Harnessing Adversarial Samples for ZSL }
\begin{algorithmic}[1]
\For{epoch $e = 0,1,\ldots,E-1$}
    \For{batch $i = 0, 1,\ldots, B-1$}
    \State Randomly select a batch $\{I, y, \phi(y) \}^{N^b}$
    \State Compute $\mathcal{L}_{CLS}$ + $\mathcal{L}_{LOC}$ {\color{blue} \# Standard ZSL training}
    \State Update model $\mW_i \leftarrow \mW_{i-1} - \eta \bigtriangledown (\mathcal{L}_{CLS} + \mathcal{L}_{LOC})$ 
        \For{$t = 0,1,\ldots,T-1$} {\color{blue} \# Perturb samples for $T$ times}
            \State Compute adversarial loss using $\mathcal{L}_{HAS}$ in Eq. \ref{eq:HAS}
            \State Update adversarial samples $I^{ad}_{t+1}$ using Eq. \ref{eq:update}
        \EndFor
        \State Compute $\mathcal{L}_{CLS}$ + $\mathcal{L}_{LOC}$ with $I^{ad}_{T}$  {\color{blue} \# Adversarial training}
        \State Update model $\mW_i \leftarrow \mW_{i-1} - \eta \bigtriangledown (\mathcal{L}_{CLS} + \mathcal{L}_{LOC})$ 
    \EndFor
\EndFor
\State Return model weights $\mW$
\end{algorithmic}

\label{alg}
\end{algorithm}
 
\subsection{Diverse Generation}
To further diversify the adversarial samples, we propose perturbing the images based on localized attributes. In attention maps, $h(x) \in \mathbb{R}^{K\times H \times W}$, each map $h(x)_k \in \mathbb{R}^{H \times W}$ corresponds to the presence of a specific attribute. By slightly shifting the presence of attributes within the images, we can achieve perturbation with respect to each individual attribute.

We can interpret the attention map as a probability distribution over regions, where the highest probability represents the attribute value. In other words, the region with the highest probability is the area to which the model exhibits the greatest sensitivity. Keeping the model fixed, an image can be altered locally as local responses change. Similar to $\mathcal{L}_{ROB}$, we employ entropy maximization to equalize the response values of different regions. Moreover, to make the sample more challenging to learn, we propose suppressing the overall attention weights by adding a regularization term to the attention maps. The diversity loss can be formulated as follows:
\begin{equation}
\begin{gathered}
\mathcal{L}_{DIV} = \Sigma_{k \in K} \lVert h(f(I^{ad}_{t}))_k \rVert_{2}^{2} - h(f(I^{ad}_{t}))_k \log h(f(I^{ad}{t}))_k,
\label{eq:att}
\end{gathered}
\end{equation}
where $h(f(I^{ad}_{t}))_k$ is the $k$-th attention map for $k$-th attribute. By incorporating localized attribute perturbation and entropy maximization, the proposed method enhances the ZSL model's ability to adapt to localized changes in the input images while maintaining robustness to adversarial perturbations.


\begin {table*}[t]
\caption {Performance comparison in accuracy (\%) of the state-of-the-art ZSL and GZSL on three datasets. For ZSL, performance results are reported with the average top-1 classification accuracy (T1). For GZSL, results are reported in terms of top-1 accuracy of unseen (U) and seen (S) classes, together with their harmonic mean (H). The best and second-best results are marked in {\color{red}{Red}} and {\color{blue}{Blue}}. APN$^{*}$ represents training with an image size of 448x448.}
\vspace{-10pt}
\centering
\scalebox{1.1}{
\begin{tabular}[t]{r|cccc|cccc|cccc}
\specialrule{.1em}{.00em}{.00em}
  \multirow{2}{*}{}      & \multicolumn{4}{c|}{CUB}  &  \multicolumn{4}{c|}{AwA2}   & \multicolumn{4}{c}{SUN}\\ 
 \cmidrule{2-13}

  \multirow{-2}{*}{Methods}  & \multicolumn{1}{c|}{\textit{T1}} &\textit{U} & \textit{S} & \textit{H}  & \multicolumn{1}{c|}{\textit{T1}} & \textit{U} & \textit{S} & \textit{H}  & \multicolumn{1}{c|}{\textit{T1}} & \textit{U} & \textit{S} & \textit{H} \\
  \specialrule{.1em}{.00em}{.00em}
           \textbf{Generative Methods}
\\
\rowcolor[gray]{.9}                    f-CLSWGAN(CVPR'18) \cite{xian2018feature}       
&\multicolumn{1}{c|}{57.3}   & 43.7           & 57.7              & 49.7
&\multicolumn{1}{c|}{65.3}   & 56.1           & 65.5              & 60.4     
&\multicolumn{1}{c|}{60.8}   & 42.6           & 36.6              & 39.4     
\\
                     CADA-VAE(CVPR'19)  \cite{schonfeld2019generalized}  
&\multicolumn{1}{c|}{60.4}    & 51.6              & 53.5              & 52.4
&\multicolumn{1}{c|}{64.0}    & 55.8              & 75.0              & 63.9
&\multicolumn{1}{c|}{61.7}    & 47.2              & 35.7              & 40.6
\\        
\rowcolor[gray]{.9}                     f-VAEGAN-D2(CVPR'19)  \cite{xian2019f}       
&\multicolumn{1}{c|}{61.0}    & 48.4              & 60.1              & 53.6
&\multicolumn{1}{c|}{71.1}    & 57.6              & 70.6              & 63.5     
&\multicolumn{1}{c|}{64.7}    & 45.1              & 38.0              & 41.3   
\\       
                    TF-VAEGAN(ECCV'20)  \cite{narayan2020latent}
&\multicolumn{1}{c|}{64.9}    & 52.8              & 64.7              & 58.1
&\multicolumn{1}{c|}{\color{blue}{72.2}}    & 59.8              & 75.1              & 66.6  
&\multicolumn{1}{c|}{\color{red}{66.0}}    & 45.6              & \color{red}{40.7}              & \color{blue}{43.0}
\\  
\rowcolor[gray]{.9}                    E-PGN(CVPR'20)  \cite{yu2020episode}
&\multicolumn{1}{c|}{72.4}    & 52.0              & 61.1              & 56.2 
&\multicolumn{1}{c|}{\color{red}{73.4}}    & 52.6              & 83.5              & 64.6    
&\multicolumn{1}{c|}{-}       & -                 & -                 & -        
\\

           SDGZSL(ICCV'21) \cite{chen2021semantics}      
&\multicolumn{1}{c|}{75.5}    & 59.9              & 66.4              & 63.0
&\multicolumn{1}{c|}{72.1}    & \color{blue}{64.6}              & 73.6              & 68.8    
&\multicolumn{1}{c|}{62.4}    & \color{blue}{48.2}              & 36.1              & 41.3    
\\
\rowcolor[gray]{.9}           HSVA(NeurIPS'21) \cite{chen2021hsva}      
&\multicolumn{1}{c|}{62.8}    & 52.7              & 58.3              & 55.3
&\multicolumn{1}{c|}{-}       & 59.3              & 76.6              & 66.8
&\multicolumn{1}{c|}{63.8}    & \color{red}{48.6} & 39.0              & \color{red}{43.3}  
\\

  \specialrule{.1em}{.00em}{.00em}
           \textbf{Embedding-based Methods}
\\
\rowcolor[gray]{.9} SP-AEN(CVPR'18) \cite{chen2018zero}
&\multicolumn{1}{c|}{55.4}    & 34.7              & 70.6              & 46.6
&\multicolumn{1}{c|}{58.5}    & 23.3              & \color{red}{90.9}              & 37.1
&\multicolumn{1}{c|}{59.2}    & 24.9              & 38.6              & 30.3   
\\
                     TCN(ICCV'19)       \cite{jiang2019transferable}   
&\multicolumn{1}{c|}{59.5}    & 52.6          & 52.0              & 52.3
&\multicolumn{1}{c|}{71.2}    & 61.2          & 65.8              & 63.4
&\multicolumn{1}{c|}{61.5}    & 31.2          & 37.3              & 34.0
\\
\rowcolor[gray]{.9}                     DVBE (CVPR'20)     \cite{min2020domain}       
&\multicolumn{1}{c|}{-}    & 53.2          & 60.2              & 56.5
&\multicolumn{1}{c|}{-}    & 63.6          & 70.8              & 67.0   
&\multicolumn{1}{c|}{-}    & 45.0          & 37.2              & 40.7
\\ 
APN(NeurIPS'20)       \cite{xu2020attribute}   
&\multicolumn{1}{c|}{72.0}    & 65.3          & 69.3              & 67.2
&\multicolumn{1}{c|}{68.4}    & 56.5          & 78.0              & 65.5
&\multicolumn{1}{c|}{61.6}    & 41.9          & 34.0              & 37.6
\\
APN$^{*}$(NeurIPS'20)       \cite{xu2020attribute}   
&\multicolumn{1}{c|}{75.6}    & 68.6          & 71.6              & 70.1
&\multicolumn{1}{c|}{69.8}    & 60.1          & 86.5              & 71.0
&\multicolumn{1}{c|}{62.6}    & 42.8          & 37.7              & 40.1
\\
\rowcolor[gray]{.9}  GEM(CVPR'21)       \cite{liu2021goal}   
&\multicolumn{1}{c|}{\color{red}{77.8}}    & 64.8          & \color{blue}{77.1}              & \color{blue}{70.4}
&\multicolumn{1}{c|}{67.3}    & \color{red}{64.8}          & 77.5              & \color{blue}{70.6}
&\multicolumn{1}{c|}{62.8}    & 38.1          & 35.7              & 36.9
\\
                       MSDN (CVPR'22)     \cite{chen2022msdn}       
&\multicolumn{1}{c|}{76.1}    & \color{red}{68.7}          & 67.5              & 68.1
&\multicolumn{1}{c|}{70.1}    & 62.0          & 74.5              & 67.7    
&\multicolumn{1}{c|}{\color{blue}{65.8}}    & 52.2          & 34.2              & {41.3} 
\\
\specialrule{.1em}{.00em}{.00em}

\rowcolor[gray]{.9}    HAS (ours)
&\multicolumn{1}{c|}{\color{blue}{76.5}} & \color{blue}{69.6} &  \color{red}{74.1} &  \color{red}{71.8}   
&\multicolumn{1}{c|}{{71.4}} & 63.1 & \color{blue}{87.3} & \color{red}{73.3}
&\multicolumn{1}{c|}{63.2} & 42.8 & 38.9 & {40.8}
\\
\specialrule{.1em}{.00em}{.00em}
\end{tabular}}
\label{gzslperoformance}
\vspace{-10pt}
\end {table*}

\subsection{Model Training and Zero-Shot Prediction}
To facilitate a clear understanding of the overall model training and zero-shot prediction process, we present the training process in Algorithm \ref{alg}. For each batch of the training data, we initally perform standard training on the model using the original data, adhering to the procedures described in Section \ref{stdZSLtraining}. Subsequently, we fix the model and compute gradients on the images for $T$ steps, generating the adversarial samples. These adversarial samples, containing attribute-conditional visual variance, are then fed into the model for ZSL training. Upon completion of the model training, the model gains ZSL capability. 

For zero-shot prediction, given a test image $I^u$ from unseen classes, we can predict the corresponding attributes and compute the compatibility score with the candidate attribute vectors. The label exhibiting the highest compatibility becomes the predicted class label:
\begin{equation}
\begin{aligned}
\hat{y^u} = \argmax_{y^u \in \mathcal{Y}^u} [g(f(I))^T \mV \phi(y^u)]
\end{aligned}
\end{equation}

For generalized zero-shot learning (GZSL), test images may originate not only from unseen classes but also from seen classes. As the model is trained exclusively on seen classes only, the prediction will inevitably exhibit bias towards seen classes. Following existing methods\cite{chao2016empirical,xu2020attribute,liu2021goal}, calibrated stacking is employed to reduce the probability score on seen classes by a calibration factor $\mu$. Consequently, the GZSL prediction is formulated as:
\begin{equation}
\begin{aligned}
\hat{y} = \argmax_{y \in \mathcal{Y}} [(g(f(I))^T \mV \phi(y)) - \mu \mathds{1}(y\in \mathcal{Y}^{s})],
\end{aligned}
\end{equation}
where $\mathds1(\cdot)$ is an indicator function that determines whether a label originates from seen or unseen classes.



\section{Experiments}

\subsection{Datasets and Implementation Details}
We conduct experiments on three widely used benchmark ZSL datasets of image classification. They are \textbf{CUB} \cite{wah2011caltech}, \textbf{AwA2} \cite{lampert2013attribute} and \textbf{SUN} \cite{patterson2014sun}.
CUB consists of 11,788 images from 200 bird species, of which 150 are selected as seen classes and 50 as unseen classes. Each class is annotated with 312 attributes. AWA2 is a considerably larger dataset with 30,475 images from 50 classes, and they are annotated with 85 attributes. SUN dataset has 14,340 images of 717 scene classes, of which 645 classes and the rest of 72 classes are chosen as seen and unseen classes, each class has 101 attributes.  

Following compared methods \cite{xu2020attribute,liu2021goal,chen2022msdn}, we adopt ResNet101 pretrained on ImageNet1k as our CNN backbone to extract feature maps. We use Adam optimizer with beta1 set to 0.5 and beta2 set to 0.999. The learning rate is initialized as 10$^4$ and decreased every ten epochs by a factor of 0.8. The batch size is set to 64. The loss weights $\lambda_1$, $\lambda_2$, $\lambda_3$ and the perturbation strength $\epsilon$ is set to between \{0.01, 1.0\}, \{0.1, 10.0\}, \{1e-5, 1e-3\} and \{1,8\}, respectively. 

To avoid the failure of classification accuracy for imbalanced class distributions, we adopt average per-class Top-1 accuracy as the fair evaluation criteria for conventional ZSL and the seen and unseen set performance in GZSL:
\begin{equation}
\begin{aligned}
  Acc_{\mathcal{Y}} = \frac{1}{|\mathcal{Y}|} \sum^{|\mathcal{Y}|}_{y=1} \frac{\# \  of \ correct \ predictions \ in \ y}{\# \ of \ samples \ in \ y},
\end{aligned}
\end{equation}
where $|\mathcal{Y}|$ is the number of testing classes. A correct prediction is defined as the highest probability of all candidate classes.
Following \cite{xian2018zero}, the harmonic mean of the average per-class Top-1 accuracies on seen $Acc_S$ and unseen $Acc_U$ classes are used to evaluate the performance of generalized zero-shot learning. It is computed by:
\begin{equation}
\begin{aligned}
  H = \frac{2*Acc_S*Acc_U}{Acc_S+Acc_U}.
\end{aligned}
\end{equation}

\begin {table}[t]
\caption {Ablation study for different components of HAS.}
\centering
\scalebox{0.85}{
\begin{tabular}[t]{l|cc|cc|cc}
\specialrule{.1em}{.00em}{.00em}
  \multirow{2}{*}{}      & \multicolumn{2}{c|}{CUB}   & \multicolumn{2}{c|}{AwA2} & \multicolumn{2}{c}{SUN}\\ 
 \cmidrule{2-7}
  \multirow{-2}{*}{Methods}  & \multicolumn{1}{c|}{\textit{T1}} & \textit{H}  & \multicolumn{1}{c|}{\textit{T1}} & \textit{H} & \multicolumn{1}{c|}{\textit{T1}} & \textit{H} 
\\
                  Baseline(APN$^*$)
&\multicolumn{1}{c|}{75.6}             & 70.1
&\multicolumn{1}{c|}{69.8}             & 71.0    
&\multicolumn{1}{c|}{62.6}             & 40.1 
\\
\rowcolor[gray]{.9}                    $+ \mathcal{L}_{CLS}$    
&\multicolumn{1}{c|}{74.3}             & 68.9
&\multicolumn{1}{c|}{68.6}             & 70.2     
&\multicolumn{1}{c|}{59.2}             & 37.8
\\
\rowcolor[gray]{.9}                   $+\mathcal{L}_{CLS}$ +  $\mathcal{L}_{ROB}$   
&\multicolumn{1}{c|}{74.7}             & 68.5
&\multicolumn{1}{c|}{68.5}             & 69.3     
&\multicolumn{1}{c|}{60.1}             & 37.9     
\\
                   $+ \mathcal{L}_{CLS}$ +  $\mathcal{L}_{DIV}$   
&\multicolumn{1}{c|}{69.2}             & 68.2
&\multicolumn{1}{c|}{67.6}             & 69.9     
&\multicolumn{1}{c|}{60.8}             & 38.4     
\\
\rowcolor[gray]{.9}                   $+\mathcal{L}_{CLS}$ + $\mathcal{L}_{ROB}$ + $\mathcal{L}_{REL}$
&\multicolumn{1}{c|}{76.0}             & 71.5
&\multicolumn{1}{c|}{71.2}             & 72.6     
&\multicolumn{1}{c|}{62.9}             & 40.3     
\\
                  $+\mathcal{L}_{CLS}$ + $\mathcal{L}_{DIV}$ + $\mathcal{L}_{REL}$
&\multicolumn{1}{c|}{76.1}             & 70.9
&\multicolumn{1}{c|}{70.7}             & 72.8     
&\multicolumn{1}{c|}{61.8}             & 40.7     
\\
\specialrule{.1em}{.00em}{.00em}

\rowcolor[gray]{.9}    HAS (ours)
&\multicolumn{1}{c|}{\textbf{76.5}}    & \textbf{71.8}   
&\multicolumn{1}{c|}{\textbf{71.4}}    & \textbf{73.3}     
&\multicolumn{1}{c|}{\textbf{63.2}}    & \textbf{40.8}
\\
\specialrule{.1em}{.00em}{.00em}
\end{tabular}}
\label{ablation}
\vspace{-10pt}
\end {table} 

\subsection{Comparison with the State-of-the-Art}
We selected recent state-of-the-art ZSL methods for comparison, which include generative methods f-CLSWGAN \cite{xian2018feature},  CADA-VAE  \cite{schonfeld2019generalized}, f-VAEGAN-D2  \cite{xian2019f}, TF-VAEGAN  \cite{narayan2020latent},  E-PGN \cite{yu2020episode}, SDGZSL \cite{chen2021semantics}, HSVA \cite{chen2021hsva}, and embedding-based methods SP-AEN \cite{chen2018zero},  TCN \cite{jiang2019transferable},  DVBE \cite{min2020domain}, APN \cite{xu2020attribute}, GEM \cite{liu2021goal} and MSDN \cite{chen2022msdn}. Table \ref{gzslperoformance} presents the performance comparison with these state-of-the-art ZSL methods on three benchmark datasets. Notably, in the original APN implementation, the image size used is 224x224. To facilitate a fair comparison with GEM and MSDN, which utilize an image size of 448x448, we have reproduced APN with the larger image size and report the performance for this modified version as APN$^{*}$. For GZSL, our methods HAS achieves the best results of 71.8\% and 73.3\% on CUB and AwA2 datasets. Compared with our baseline method APN, we achieve consistent performance improvement on all three datasets and both conventional and generalized settings, which demonstrates the effectiveness of adversarial samples. As the SUN dataset involves a substantial number of classes, \textit{i.e.,} 717, generative methods can better generalize to unseen classes than embedding-based methods. However, we still achieve comparable performance with other embedding-based methods. 

\begin{figure*}[t]
	\centering
	\subfloat[Reliability weight $\lambda_1$]
	{\includegraphics[width=1.7in]{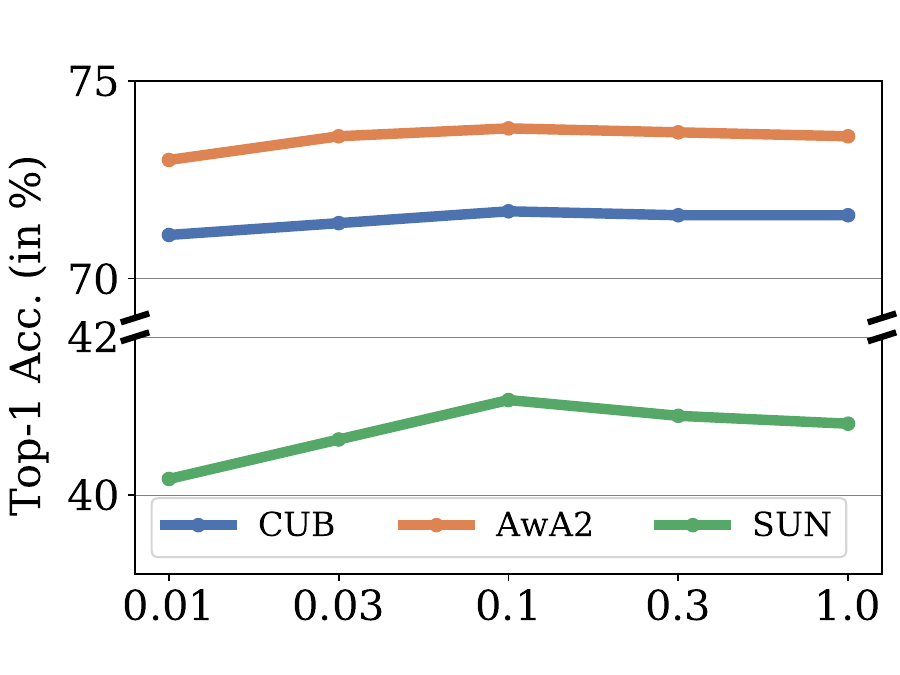}}
 	\subfloat[Robustness weight $\lambda_2$]
	{\includegraphics[width=1.7in]{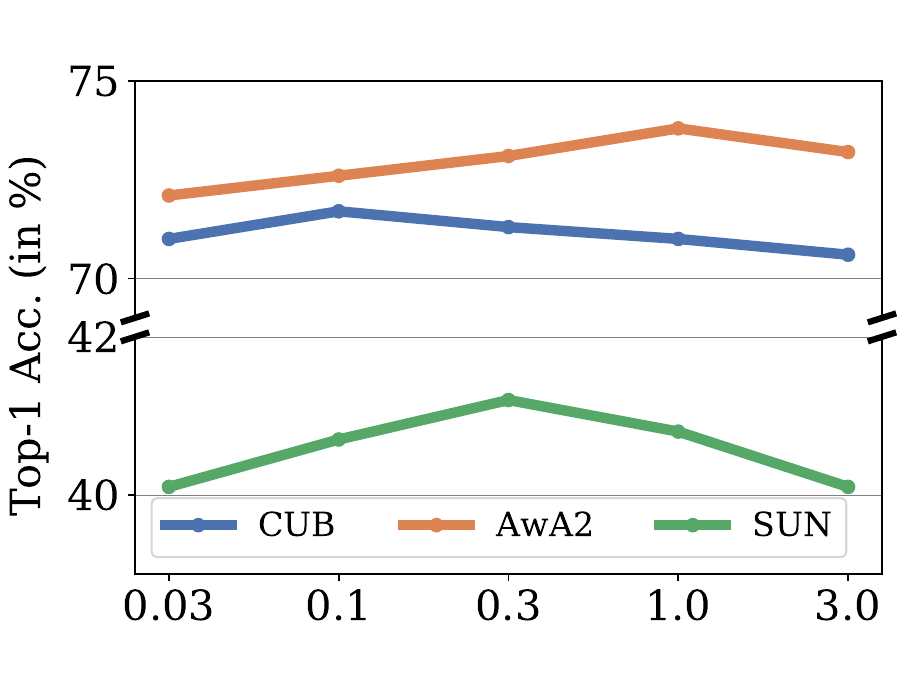}}
	\subfloat[Diversity weight $\lambda_3$]
	{\includegraphics[width=1.7in]{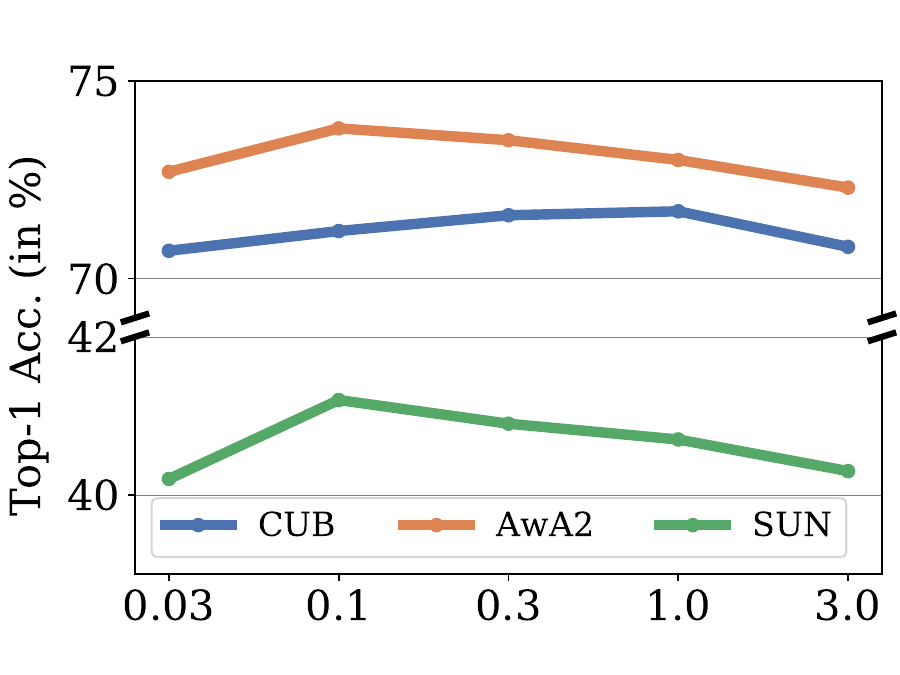}}
	\subfloat[Perturbation strength $\epsilon$]
	{\includegraphics[width=1.7in]{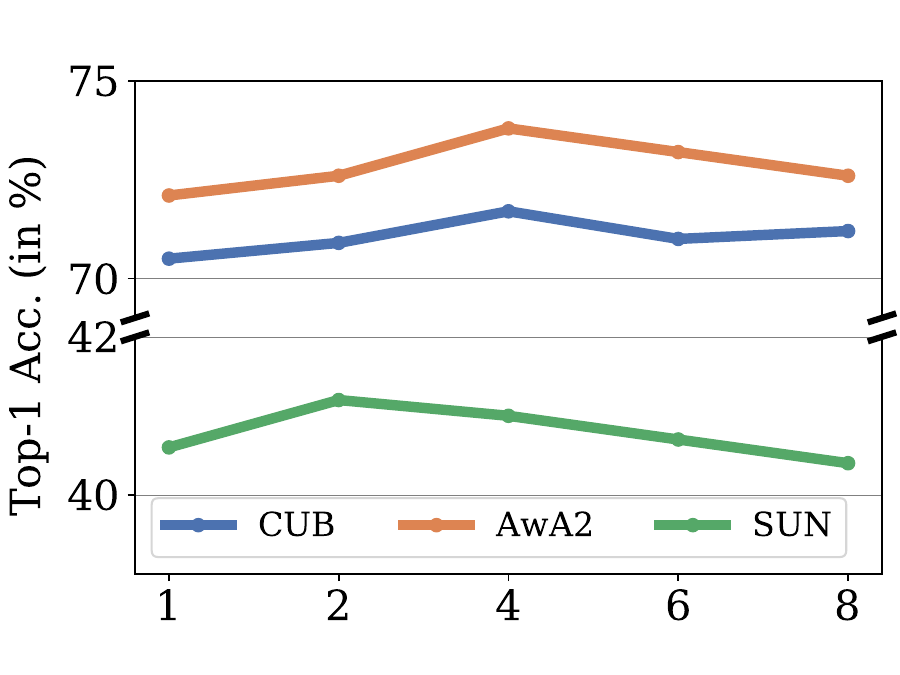}}\\
	 \vspace{-5pt}
  \caption{Hyper-parameter sensitivity. The horizontal axis indicates the varying hyper-parameters for (a) Reliability weight $\lambda_1$, (b) Robustness weight $\lambda_2$, (c) Diversity weight $\lambda_3$ and (d) Perturbation strength $\epsilon$.}\label{hyper}
 \vspace{-10pt}
\end{figure*}

\subsection{Ablation Study}
\begin{figure}[t]
	\centering
	\subfloat[w/o $\mathcal{L}_{REL}$]
	{\includegraphics[width=1.6in]{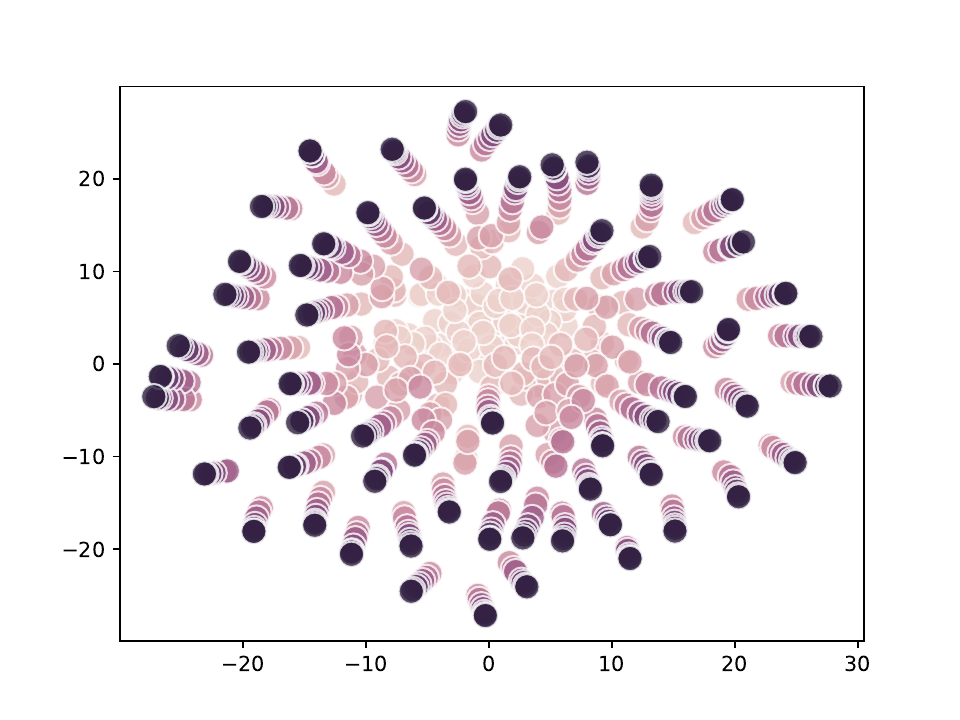}}
	\subfloat[w/ $\mathcal{L}_{REL}$]
	{\includegraphics[width=1.6in]{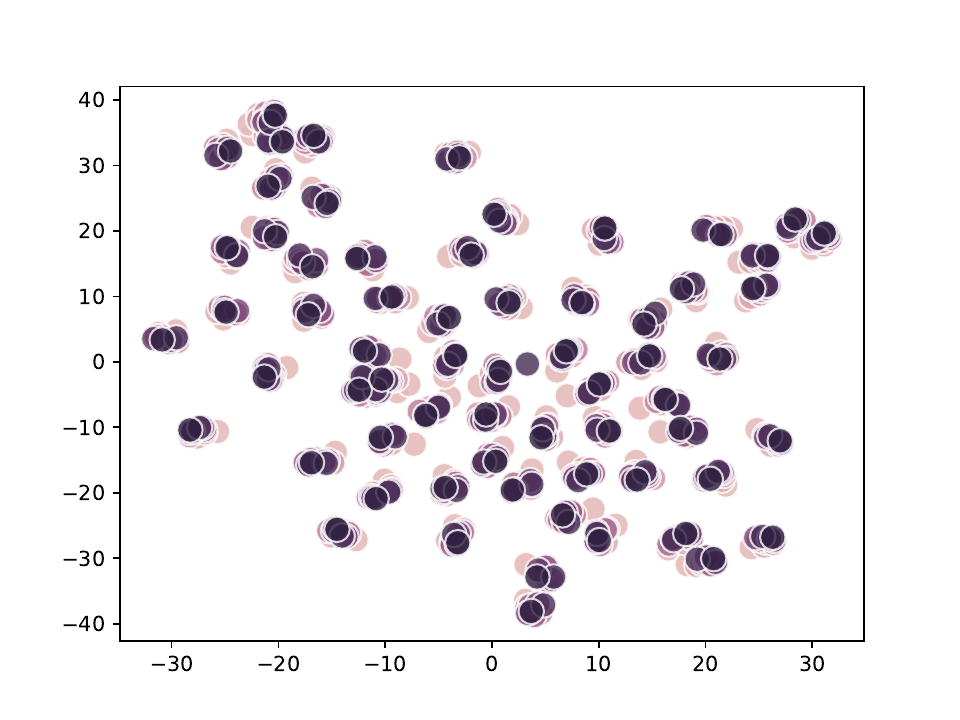}}
  \vspace{-5pt}
	\caption{Visualization of the visual features. Darker colors represent visual features in the later perturbation stages. (a) The drastic changes without constraint to visual features lead to semantic distortion. (b) $\mathcal{L}_{REL}$ avoids the visual features shifting from the original positions.  }\label{prec_cifar}
 \vspace{-15pt}
 \label{mea_vis}
\end{figure}

To analyze the contribution of each perturbation component, we conduct an ablation study on the complete adversarial training strategy. 
As shown in Table \ref{ablation}, we decompose our complete training strategy into seven different combinations. 
Baseline(APN$^*$) is the standard ZSL model APN trained with the image size of 448×448 without adversarial training. In $+\mathcal{L}_{CLS}$, we incorporated adversarial training and generate adversarial samples with the classification loss only. This variant slightly worsens the ZSL performance on the three datasets, which confirms the conclusion drawn from \cite{DBLP:conf/iclr/TsiprasSETM19,raghunathan2020understanding} the robustness and accuracy tradeoff in the vanilla adversarial training. In both $+\mathcal{L}_{CLS}+\mathcal{L}_{ROB}$ and $+\mathcal{L}_{CLS}+\mathcal{L}_{DIV}$ , the robustness loss cannot improve the performance. This confirms the utility of the reliability loss $\mathcal{L}_{MEA}$, which is necessary to prevent semantic distortion when perturbing the images. We also visualize the difference of visual features with and without the reliability loss $\mathcal{L}_{REL}$ as shown in Figure \ref{mea_vis}. We perturb 64 random images for 10 times, darker colors represent later perturbation stages. The visual features significantly shift from the original space without the constraint loss, while this issue is effectively mitigated when applying $\mathcal{L}_{REL}$. We also report the performance of the two variants without $\mathcal{L}_{DIV}$ or $\mathcal{L}_{ROB}$, both of which can achieve performance increase on the baseline method. When combining all the components, we achieve the best performance.



\subsection{Hyper-parameter Sensitivity}
In a series of experiments, we investigate the influence of the four hyper-parameters on the performance of our proposed method. Figure \ref{hyper} (a) shows the effect of varying the constraint weight $\lambda_1$ from 0.01 to 1.0. As the reliability weight increases, the overall performance also improves. However, the performance saturation occurs at around 0.1, and larger reliability weights tend to decrease performance due to over-controlled perturbation. Figure \ref{hyper} (b) demonstrates that the robustness weight is a considerably sensitive hyper-parameter. We observe that coarse-grained datasets tend to perform better when the robustness weight is high. In Figure \ref{hyper} (c), we vary the diversity weight $\lambda_3$ and found that AwA2 and SUN both achieve the best performance at 0.1, while CUB peaks at 1.0. The perturbation strength can be considered the learning rate of generating adversarial samples. In Figure \ref{hyper} (d), CUB and AwA2 achieve the best performance at 4, while SUN peaks at 2.

\begin {table}[t]
\caption {Effects of different traditional augmentation techniques on the CUB dataset.}
\vspace{-10pt}
\begin{center}
\scalebox{0.87}{
\begin{tabular}[t]{l m{0.7cm} m{0.7cm} | m{0.7cm}m{0.7cm} | m{0.7cm}m{0.7cm}}\toprule
& \multicolumn{2}{c|}{CUB}  & \multicolumn{2}{c|}{AwA2}  &  \multicolumn{2}{c}{SUN}\\ 
\cmidrule(lr){2-3} \cmidrule(lr){4-5} \cmidrule(lr){6-7}   
& CZSL  &  GZSL  & CZSL  &  GZSL  & CZSL  &  GZSL              \\
\noalign{\smallskip}\hline\noalign{\smallskip}
{Baseline}  
    & 72.0  & 67.2  & 68.4  & 67.4  & 61.6  & 37.6             \\ 
\noalign{\smallskip}\hline\noalign{\smallskip}
{{ColorJitter0.2}}
    & 66.8  & 61.1  & 66.5  & 65.1  & 59.9  & 37.4             \\
{{ColorJitter0.4}}
    & 67.3  & 60.9  & 66.6  & 64.9  & 59.7  & 37.0             \\
{{GrayScale0.2}}
    & 69.6  & 65.1   & 68.0 & 66.8  & 60.1  & 37.1             \\
{{GrayScale0.4}}
    & 68.4  & 63.3  & 66.6  & 66.0  & 59.9  & 37.0             \\
\noalign{\smallskip}\hline\noalign{\smallskip}
{{GaussianBlur(L)}}
    & 71.1  & 65.9  & 69.1  & 65.7  & 60.8  & 37.0             \\
{{GaussianBlur(H)}}
    & 68.7  & 62.1  & 62.5  & 56.2  & 58.5  & 34.3             \\
\noalign{\smallskip}\hline\noalign{\smallskip}
{{RandomRotate}}
    & 65.8  & 59.5  & 56.6  & 54.8  & 46.7  & 21.7             \\
{{RandomCrop}}
    & 67.0  & 60.8  & 60.8  & 58.8  & 36.1  & 14.9             \\
\noalign{\smallskip}\hline\noalign{\smallskip}
{{CutOut}}
    & 70.5  & 65.5  & 62.5  & 60.2  & 60.1  & 35.2             \\
{{MixUp}}
    & 66.4  & 59.5  & 38.9  & 39.1  & 51.8  & 28.2             \\
{{CutMix}}
    & 68.5  & 62.4  & 53.6  & 42.0  & 57.2  & 29.8             \\
{{SnapMix}}
    & 69.0  & 61.1 & 51.3   & 45.4  & 56.5  & 30.1             \\
\noalign{\smallskip}\hline\bottomrule
\end{tabular}}
\end{center}
\label{tab:aug}
\vspace{-15pt}
\end {table} 

\subsection{Traditional Image Augmentation Results}

We explore a variety of representative image augmentation techniques in the training of the ZSL model APN, including Color Jitter, Grayscale, Gaussian Blurring, random rotation and random crop. For each augmentation, we apply both mild and strong augmentation to see their effects. For random rotation, the rotation degree ranges from 0 and 360 degrees. For random crop, we crop half the size of the original images. As reported in Table \ref{tab:aug}, there is a consistent performance drop across all datasets and in both conventional zero-shot learning and generalized zero-shot learning settings.
Notably, we take one step further by exploring state-of-the-art data mixing augmentation strategies, going beyond traditional techniques. We examine methods such as CutOut \cite{DBLP:journals/corr/abs-1708-04552}, MixUp \cite{DBLP:conf/iclr/ZhangCDL18}, CutMix \cite{DBLP:conf/iccv/YunHCOYC19}, and SnapMix \cite{huang2021snapmix}. These approaches have demonstrated their effectiveness in enhancing the performance of deep neural networks. In contrast to traditional image augmentation techniques that solely modify an image's appearance, these methods alter the underlying pixel values of the image.
CutOut augments data by removing a rectangular region from an image. MixUp combines images and blends labels using linear combinations. CutMix employs a cut-and-paste operation to mix images, and it combines labels based on the area ratio. SnapMix estimates the semantic structure of a synthetic image by leveraging class activation maps. However, unfortunately, semantic space is usually not  on images also cause inconsistency between images and attributes, \textit{i.e.,} causing semantic distortion.


\begin{figure*}[t]
    \centering
    \includegraphics[width=0.95\linewidth]{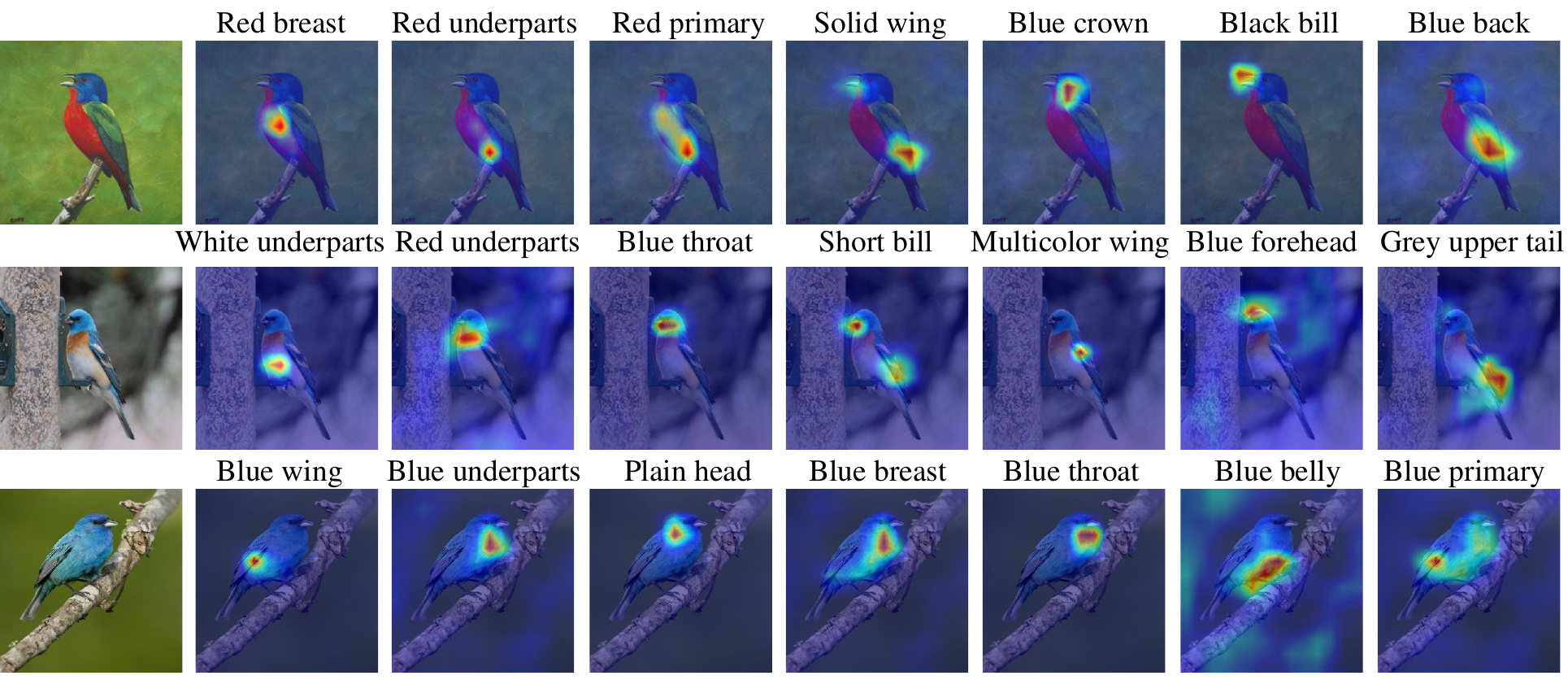}
    \vspace{-10pt}
    \caption{Visualization of the predicted attribute attention maps on CUB dataset.}
    \vspace{-5pt}
    \label{fig:att_vis}
\end{figure*}

\begin{figure}[t]
    \centering
    {\includegraphics[width=0.62in]{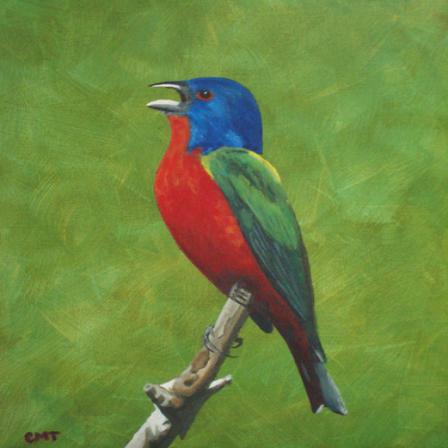}}\hspace{-2.5pt}
    {\includegraphics[width=0.62in]{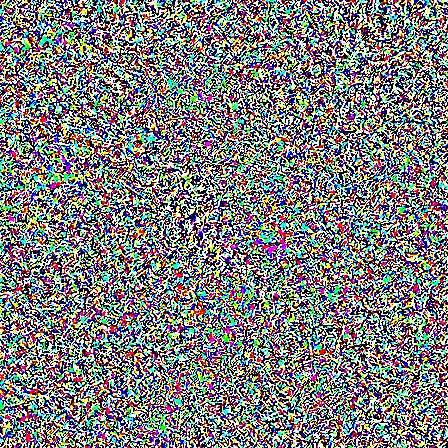}}\hspace{-2.5pt}
    {\includegraphics[width=0.62in]{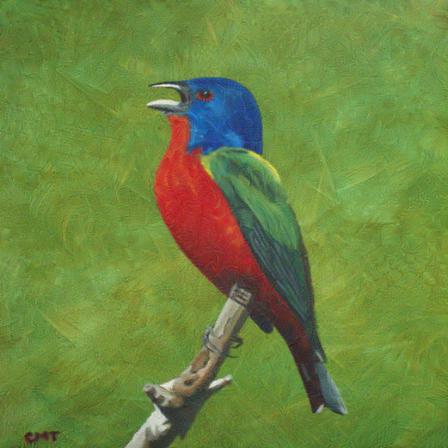}}\hspace{-2.5pt}
    {\includegraphics[width=0.62in]{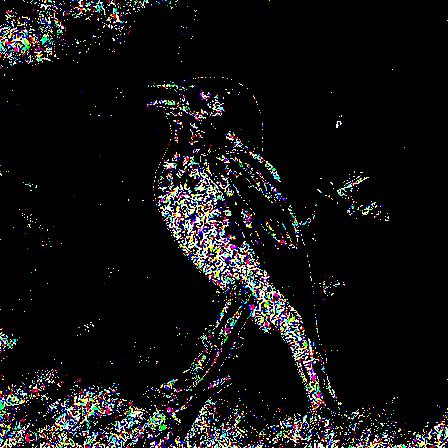}}\hspace{-2.5pt}
    {\includegraphics[width=0.62in]{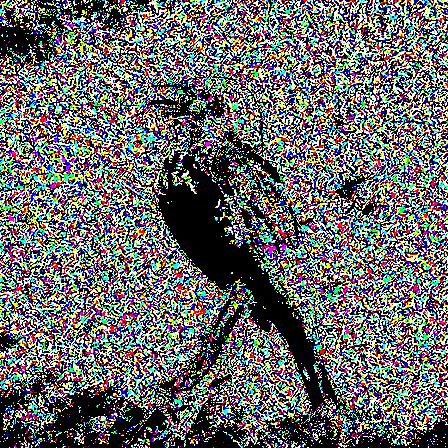}}\\
    \subfloat[]
    {\includegraphics[width=0.62in]{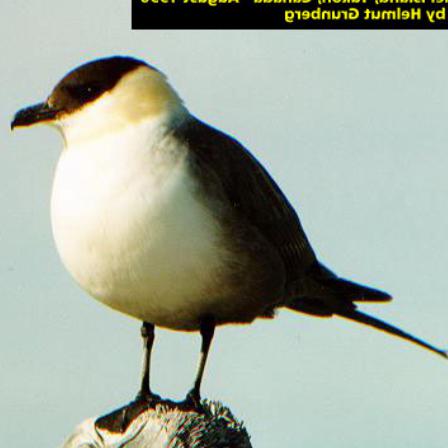}}
    \subfloat[]
    {\includegraphics[width=0.62in]{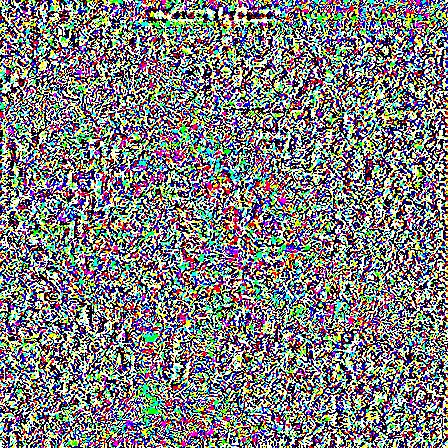}}
    \subfloat[]
    {\includegraphics[width=0.62in]{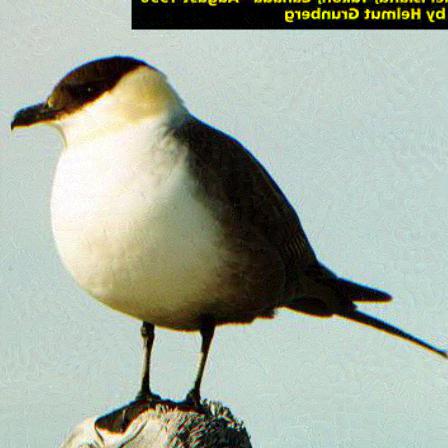}}
    \subfloat[]
    {\includegraphics[width=0.62in]{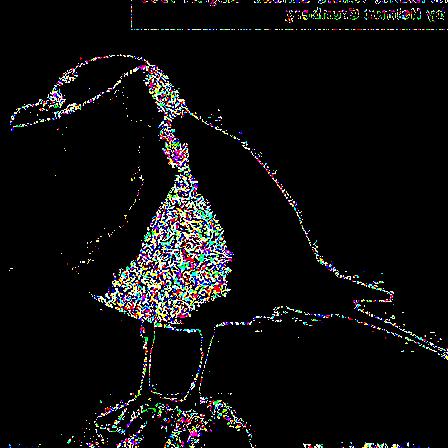}}
    \subfloat[]
    {\includegraphics[width=0.62in]{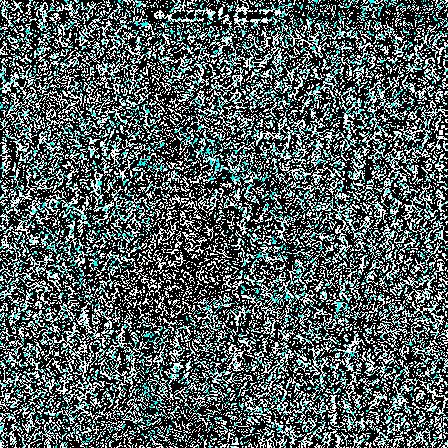}}\\
        \vspace{-10pt}
        \caption{Visualization of the perturbation components. (a) the original images; (b) the learned overall perturbation; (c) the adversarial samples after perturbation; (d) the foreground perturbation after striping the background perturbation; (e) the perturbation on the background.}\label{perturbation_vis}
        \vspace{-10pt}
        \label{perturvis}
\end{figure}

\subsection{Qualitative Results}
\subsubsection{Perturbation Visualization}

Perturbations on images are often imperceptibly small, making it challenging to distinguish augmentations from the original images. Nonetheless, it is important to note that even such seemingly negligible perturbations can significantly influence the model's behavior \cite{DBLP:conf/iclr/TsiprasSETM19,raghunathan2020understanding}. This counterintuitive phenomenon can be attributed to the high-dimensional space of the data and the intricacies of non-linear computations performed by deep learning models.
To gain a deeper understanding of adversarial training behavior, we normalize the perturbations applied to the images, i.e., the differences between the augmentations and the original images. Figure \ref{perturvis} illustrates this: (a) displays the original two images, (c) shows the adversarial samples, and (b) presents the normalized differences. By zooming in, subtle differences become more visible.
We empirically discover that the perturbation background is dominated by a few similar values, as demonstrated in (e). By removing these background values, we can isolate the foreground perturbations, as shown in (d). This observation confirms that our attribute-conditional adversarial training can effectively perceive the differences between the background and the foreground objects.

\subsubsection{Attribute Attention Visualization}
We present attention maps for randomly selected attributes on three example images from the CUB dataset. The attention maps are generated by applying max pooling operations to the model output, highlighting the regions where the model is focusing on each attribute. As our local perturbation component specifically focuses on changing the attended areas that ZSL model paying most attention to, which prevents overfitting to the easily attended areas. This visualization in Figure \ref{fig:att_vis} confirms that our method localization ability for each attribute, demonstrating its capacity to recognize and differentiate the specific features within the images. This visualization helps to better understand the model performance and interpretability in the context of zero-shot learning. Furthermore, the attention maps reveal that the attended areas are relatively small due to the local perturbation constraint, which fosters more precise localization. 
This property contributes to the model overall effectiveness in handling zero-shot learning tasks. 



\section{Conclusion}
In this paper, we propose a novel approach to harness the adversarial samples for enhancing the generalization ability of zero-shot learning models. By addressing the challenges of generating adversarial samples for ZSL, we have developed a method that incorporates three key learning properties, enabling effective utilization of adversarial samples to boost ZSL performance. The proposed approach demonstrates the potential of adversarial samples in improving the generalization capabilities of ZSL models, while mitigating the semantic distortion issues inherent in traditional image augmentation techniques.
We hope our experimental study will help understand the difference in model behavior between single-label supervision and semantic attributes supervision, and pave the way for developing more robust semantic-condition visual augmentations.

\begin{acks}
This work was partially supported by Australian Research Council CE200100025, DE200101610.
\end{acks}

\balance

\bibliographystyle{ACM-Reference-Format}
\bibliography{main}

\end{document}

%% file: math.tex
\usepackage{amsmath,amsfonts,bm}

\def\eqref#1{equation~\ref{#1}}

\def\1{\bm{1}}

\def\vx{{\bm{x}}}

\def\mV{{\bm{V}}}
\def\mW{{\bm{W}}}

\DeclareMathAlphabet{\mathsfit}{\encodingdefault}{\sfdefault}{m}{sl}
\SetMathAlphabet{\mathsfit}{bold}{\encodingdefault}{\sfdefault}{bx}{n}

\DeclareMathOperator*{\argmax}{arg\,max}

\DeclareMathOperator{\sign}{sign}